%% file: paper.tex
\documentclass[10pt,conference,a4paper]{IEEEtran}
\ifCLASSINFOpdf
  \usepackage[pdftex]{graphicx}
\else
  \usepackage[dvips]{graphicx}
\fi

\usepackage{hyperref}

\newcommand{\dquote}[1]{``#1''}

\begin{document}
%
\title{A Novel Region of Interest Extraction Layer\\for Instance Segmentation}

\author{\IEEEauthorblockN{Leonardo Rossi}
\IEEEauthorblockA{IMP Lab - D.I.A.\\
University of Parma\\
Parma, Italy\\
Email: leonardo.rossi@unipr.it}
\and
\IEEEauthorblockN{Akbar Karimi}
\IEEEauthorblockA{IMP Lab - D.I.A.\\
University of Parma\\
Parma, Italy\\
Email: akbar.karimi@unipr.it}
\and
\IEEEauthorblockN{Andrea Prati}
\IEEEauthorblockA{IMP Lab - D.I.A.\\
University of Parma\\
Parma, Italy\\
Email: andrea.prati@unipr.it}
}


%


\maketitle

\input{abstract}

%
\IEEEpeerreviewmaketitle

\input{introduction}
\input{related}
\input{groi}
\input{experiments}
\input{experiments_sota}
\input{conclusion}

\section*{Acknowledgment}

This research benefits from the HPC (High Performance Computing) facility of
the University of Parma, Italy.



\bibliographystyle{IEEEtran}
\bibliography{egbib}
%
%
%

\end{document}

%% file: abstract.tex
\begin{abstract}
Given the wide diffusion of deep neural network architectures for computer vision tasks, several new applications are nowadays more and more feasible. Among them, a particular attention has been recently given to instance segmentation, by exploiting the results achievable by two-stage networks (such as Mask R-CNN or Faster R-CNN), derived from R-CNN. In these complex architectures, a crucial role is played by the Region of Interest (RoI) extraction layer, devoted to extracting a coherent subset of features from a single Feature Pyramid Network (FPN) layer attached on top of a backbone.

This paper is motivated by the need to overcome the limitations of existing RoI extractors which select only one (the best) layer from FPN. Our intuition is that all the layers of FPN retain useful information. Therefore, the proposed layer (called Generic RoI Extractor - GRoIE) introduces non-local building blocks and attention mechanisms to boost the performance.

A comprehensive ablation study at component level is conducted to find the best set of algorithms and parameters for the GRoIE layer. 
Moreover, GRoIE can be integrated seamlessly with every two-stage architecture for both object detection and instance segmentation tasks. Therefore, the improvements brought about by the use of GRoIE in different state-of-the-art architectures are also evaluated. 
The proposed layer leads up to gain a 1.1\% AP improvement on bounding box detection and 1.7\% AP improvement on instance segmentation.

The code is publicly available on GitHub repository at \url{https://github.com/IMPLabUniPr/mmdetection/tree/groie_dev}
\end{abstract}

%% file: introduction.tex
\section{Introduction}

\begin{figure*}[thb]
	\begin{center}
		\includegraphics[width=1\textwidth]{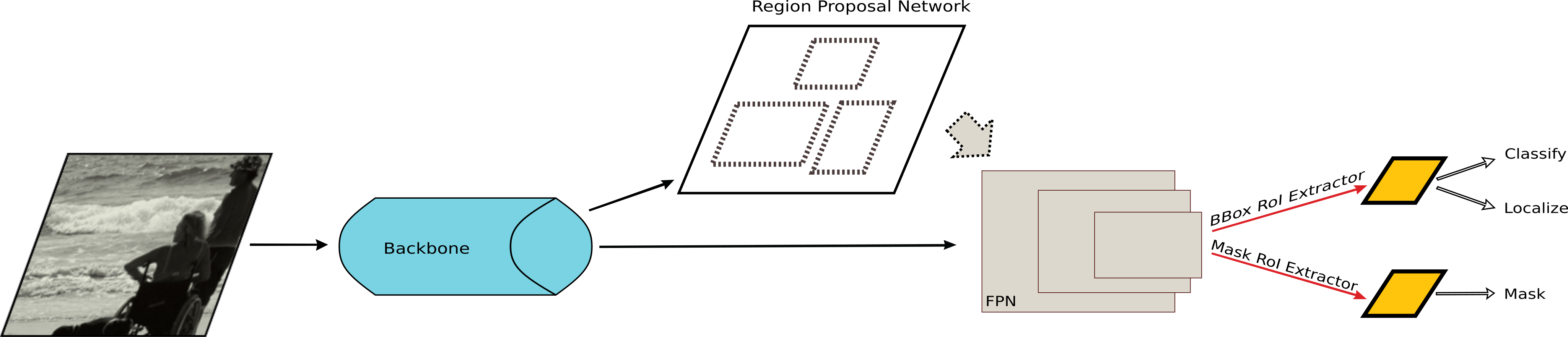}
	\end{center}
	\caption{Typical components of a two-stage R-CNN-like architecture for instance segmentation.}
	\label{fig:twostep}
\end{figure*}

Nowadays, instance segmentation is one of the most studied topics in the computer vision community. It differs from both object detection, where the final output is the set of rectangular bounding boxes which localize and classify any object instance, and semantic segmentation, where the goal is to classify any image pixel without considering if it is part of a specific instance.
In instance segmentation, the final goal is to be able to cut the single instances of objects from the original image. Its characteristics make this task very useful for several advanced applications, such as object relationship detection, automatic image captioning, content-based image retrieval, and many others.

In the recent literature, many studies have addressed the instance segmentation problem. The proposed architectures can be grouped into two main categories: one-step and two-step architectures.
The one-step architectures obtain the results with a single pass, making a direct prediction from the input image. On the contrary, an architecture belonging to the second category (two-step) is usually composed of a Region Proposal Network (RPN) \cite{girshick2014rich}, which returns a list of Regions of Interest (RoI) that are likely to contain the searched object, followed by a more specialized network with the purpose of detecting or segmenting the object/instance within each of the bounding boxes found. These networks descend from their ancestor network called R-CNN \cite{girshick2014rich}.

The typical components of a two-step architecture are shown in Fig. \ref{fig:twostep}.
As it can be seen in the diagram, the layer (highlighted in red) connecting the two steps is usually represented by the RoI extractor, which is the main focus of this paper.
Since this layer plays a crucial role in terms of final results, it should be carefully designed to minimize the loss of information.

The main objective of this layer is to perform pooling in order to transform the input region, which can be of any size, to a fixed-size feature map.
Several previous papers have tackled this problem using different RoI pooling algorithms such as RoI Align \cite{he2017mask}, RoI Warp \cite{dai2016instance} and Precise RoI Pooling \cite{jiang2018acquisition}.
Since instances of objects can appear in the image with different scales, the existing architectures (as shown in Fig. \ref{fig:twostep}) exploit a Feature Pyramid Network (FPN) \cite{lin2017feature} combined with an RPN  (e.g. Fast R-CNN \cite{girshick2015fast}, Faster R-CNN \cite{ren2015faster} and Mask R-CNN \cite{he2017mask}), to generate multi-scale feature maps.
An FPN is composed of a bottom-up pathway, where final convolutional layers from the backbone are often chosen, followed by a top-down pathway to reconstruct spatial resolution from the upper layers of the pyramid that have a higher semantic value.
With the introduction of a FPN, the fundamental issue is the selection of a FPN layer to which the RoI pooler will be applied.

Traditional methods make the selection based on the RoI obtained by the RPN.
They use the formula proposed by \cite{lin2017feature} to discover the best $\emph{k-th}$ layer to sample from, which is based on the width $w$ and height $h$ of the RoI as follows:
\begin{equation}
 k=\left\lfloor k_0 + \log_2\left(\sqrt{wh}/244\right)\right\rfloor
\end{equation}
\noindent where $k_0$ represents the highest level feature map and $224$ is the typical image size used to pre-train the backbone with ImageNet dataset.
This hard selection of a single layer of FPN might limit the power of the network's description and our intuition (supported by previous works, such as \cite{liu2018path}) is that if all scale-specific features are retained, better object detection and segmentation results can be achieved.

The main contributions of this paper are the following:
\begin{enumerate}
\item A novel RoI extraction layer called GRoIE is proposed, with the aim of a more generic, configurable and interchangeable framework for RoI extraction in two-step architectures for instance segmentation.
\item Exhaustive ablation study on different components of the proposed layer is conducted in order to evaluate how the performance changes depending on the various choices.
\item GRoIE is introduced to the major state-of-the-art architectures to demonstrate its superior performance with respect to traditional RoI extraction layers.
\end{enumerate}

The paper is organized as follows.
Section \ref{sec:related} describes the state of the art.
In Section \ref{generic_roi_extractor}, the proposed architecture is described in detail. Section \ref{sec:experiments} describes the experimental methodology as well as our in-depth ablation study on component selection. Additionally, in this section, we show how the inclusion of GRoIE layer in state-of-the-art architectures can lead to significant improvements in the overall performance.

%% file: related.tex
\section{Related Work}\label{sec:related}

As mentioned in the introduction, modern detectors employ a RoI extraction layer to select the features produced by the backbone network according to the candidate bounding boxes coming from a RPN.
This layer was first introduced in R-CNN network. Since then, many architectures derived from R-CNN (e.g., Mask R-CNN, Grid R-CNN \cite{lu2019grid}, Cascade R-CNN \cite{cai2019cascade}, HTC \cite{chen2019hybrid} and GC-net \cite{cao2019gcnet}) have used this layer as well. 
Usually, to be more invariant to object scale, the layer is not directly applied to the backbone features, but instead to an FPN attached on top of the backbone.

In \cite{lin2017feature}, a RoI pooling action is applied to a single heuristically-selected FPN output layer. This approach suffers from a problem related to untapped information.
In \cite{pont2016multiscale}, the authors propose to extract mask proposals from each scale separately, rescale them and include the resulting scales in a unique multi-scale ranked list. Eventually, only the best proposals are selected.
In \cite{ren2016object}, the authors propose to fuse features belonging to different scales by max function, using an independent backbone for each image scale. In our work, on the contrary, we utilize a feature pyramid to simplify the network and avoid doubling the number of parameters for each scale.
In SharpMask \cite{pinheiro2016learning}, the authors make a coarse mask prediction after which they fuse feature layer back in a top-down fashion until reaching the same size of the input image.
In PANet \cite{liu2018path}, the authors highlight that the information is not strictly connected with a single layer of the FPN.
They propagate low-level features, building another FPN-like structure coupled with the original FPN, where the RoI-pooled images are combined.
Our proposed GRoIE layer is inspired by this approach with the difference that it is more lightweight because of not using any extra FPN-coupled stack and proposes a novel way to aggregate data from the RoI-pooled features.
Auto-FPN \cite{xu2019auto} extends PANet model by applying the Neural Architecture Search (NAS) concept.
Also, AugFPN \cite{guo2019augfpn} can be considered an extension of PANet model.
The module we directly compare our module with is the Soft RoI Selector, which performs a RoI pooling on each FPN layer for concatenating the results.
Subsequently, through the \emph{Adaptive Spatial Fusion}, they are combined to create a weight map which passes through 1x1 and 3x3 convolutions sequentially.
In our case, we first apply a distinct convolutional operation on each layer of the FPN output which very effectively helps the network to automatically focus on the best scales.
Next, we apply a sum instead of concatenation because we have proven it has a greater learning potential for the network.
Finally, an attention layer is applied that combines fully-connected layers and convolutions to further filter the multi-scale context.

In Multi-Scale Subnet \cite{linh2018multi}, authors propose an alternative method to RoI Align which uses crop-resized branches to extract the RoI at different scales.
They use convolution with 1x1 kernel to simply maintain the same number of outputs for each branch without the purpose of helping the network to process data.
Then, before summing all branches, they apply an average pooling to reduce each branch to the same size.
Finally, a convolutional layer with 3x3 kernel is used as post-processing stage.
In our ablation study, we demonstrate that these convolutional configurations for pre- and post-processing are not the best ones possible to achieve better performance.

IONet \cite{bell2016inside} proposes not to use any FPN network but concatenated, re-scaled and dimension-reduced features directly from the backbone before performing classification and bounding box regression.
Finally, Hypercolumn \cite{hariharan2015hypercolumns} employs a hypercolumn representation to classify a pixel, using convolutions with 1x1 kernel and up-sampling the results to a common size to be able to sum them all.
In this case, the absence of a optimized RoI pooling solution and an FPN can negatively affect the final performance.
Moreover, simply processing columns of pixels taken from different stages of the backbone can be a limitation.
In fact, in our ablation study we will demonstrate that adjacent pixels are important for optimally extracting information within the various features.

%% file: groi.tex
\section{Generic RoI Extraction Layer}
\label{generic_roi_extractor}

The FPN is an architecture commonly used to extract features from different image resolutions. It has been demonstrated to have an effective power to maintain spatial information avoiding the expensive computation caused by a separate elaboration of each scale.
Inside a two-stage detection framework, one FPN output layer is heuristically selected as unique source of RoI Pooling action.
Although the formula is well thought out, it is clear that the layer selection is the result of an arbitrary choice.

\begin{figure}[b!]
	\begin{center}
		\includegraphics[width=1\linewidth]{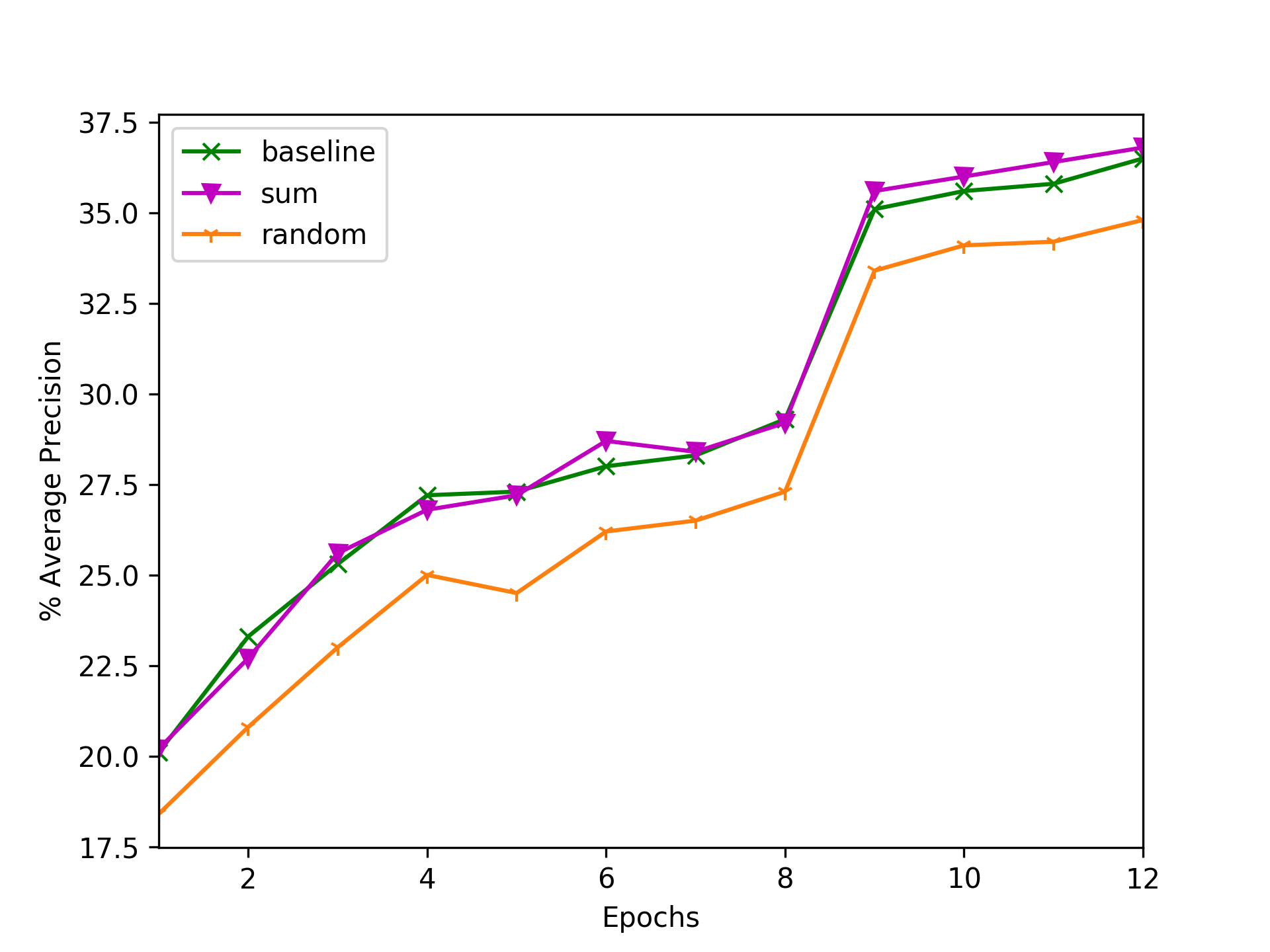}
	\end{center}
	\caption{Average precision trend for different FPN layer selection strategies. Training and testing are performed on COCO minival dataset with 12 training epochs.}
	\label{fig:groi_random}
\end{figure}

In order to demonstrate this statement, we have compared this heuristic (proposed by \cite{lin2017feature}) as baseline with a random selection of the FPN layer to sample from.
Table \ref{groi-compare-random-select} shows the average precision (AP) with different metrics (detailed in Section \ref{subsec:dataset}).
Comparing the first two rows of the table, it is evident that the difference between the randomly-selected and the heuristic choice is not enormous.
As a further proof, Fig. \ref{fig:groi_random} shows the progress with training epochs and demonstrates that the progress is similar.
This is understandable considering that each FPN layer is derived from the previous one.
It means that information is existent in the FPN layers, but in a more or less tangled way to be classified by the following modules of the network.

These results highlight that the network is capable of extracting information with good enough quality to discriminate classes from any available scale. To corroborate this finding, we have also tried to sum the FPN layers, obtaining an improvement of 0.3\% in average precision (see Table \ref{groi-compare-random-select} and Fig. \ref{fig:groi_random}).
This enhancement suggests that if all the layers are aggregated appropriately, it is more likely to produce higher quality features.

\begin{table}
	\begin{center}
		\begin{tabular}{l|c|c|c|c|c|c}
			Method & $AP$ & $AP_{50}$ & $AP_{75}$	& $AP_{s}$ & $AP_{m}$ & $AP_{l}$ \\
			\hline\hline
			baseline \cite{lin2017feature} & 36.5 & 58.4 & 39.1 & 21.9 & 40.4 & 46.8 \\
			random    & 34.8 & 56.9 & 37.0 & 19.1 & 39.3 & 45.2 \\
			sum       & \textbf{36.8} & \textbf{59.0} & \textbf{39.5} & \textbf{22.0} & \textbf{41.0} & \textbf{47.2} \\
		\end{tabular}
	\end{center}
	\caption{Comparison of different methods for selecting FPN layers. Training and testing are performed on COCO minival dataset with 12 training epochs. For explanation of the different evaluation metrics in the table columns, please refer to section \ref{subsec:dataset}}
	\label{groi-compare-random-select}
\end{table}

\begin{figure*}[thb]
	\begin{center}
	\includegraphics[width=0.8\linewidth]{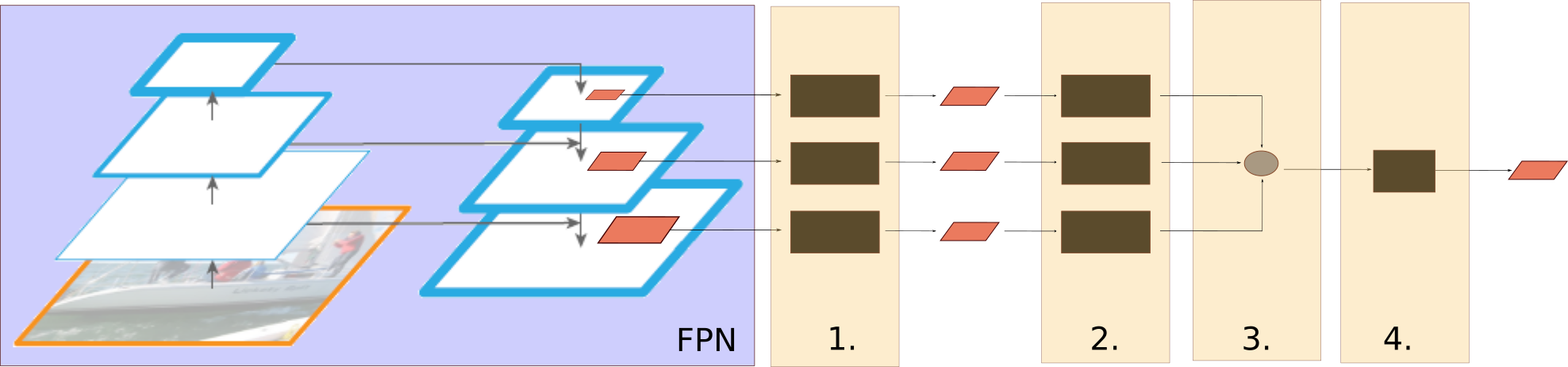}
	\end{center}
	\caption{Generic RoI Extraction framework. (1) RoI Pooler. (2) Preprocessing phase. (3) Aggregation function. (4) Post-processing phase.}
	\label{fig:groi}
\end{figure*}

Based on these preliminary ideas, we propose a novel RoI extraction layer called Generic RoI Extractor (\emph{GRoIE}) whose architecture can be seen in Fig. \ref{fig:groi}.

GRoIE is composed of the following modules:

\begin{enumerate}
	\item \textbf{RoI pooler module}:
	it is a module that performs a max pooling on non-uniform region of interest to obtain a fixed-size representation.
	Currently, many pooling techniques such as RoI Pooling \cite{girshick2015fast} and RoI Align \cite{he2017mask} are available.
	Among the existing RoI pooling techniques, we found RoI Align \cite{he2017mask} as the most appropriate since it reduces a rectangular feature map region by dividing the original RoI in equal boxes and applying bilinear interpolation inside each of them. This helps to avoid pixel quantization.
	\item \textbf{Pre-processing module}:
	its objective is to apply a preliminary elaboration to the pooled regions.
	This gives the network an additional degree of freedom which is specific for each image scale.
	This module is devoted to pre-processing the feature maps and it is usually obtained by means of a convolutional layer associated with each image scale.
	As will be shown in the ablation analysis reported in Section \ref{Component ablation study}, the optimal configuration consists of a single 5x5 convolutional layer per scale.
	Our experiments suggest that it is not convenient to process the features individually which can be explained by acknowledging that each feature is semantically connected with adjacent features. This is particularly true, remembering that the final objective is object detection/segmentation and, usually, objects are spread over a consistent region of the image.
	\item \textbf{Aggregation module}:
	it defines how to aggregate the single RoIs coming from each branch.
	The most frequent operations are concatenation and summation.
	There are multiple ways of merging different branches.
	After our ablation analysis, we found that the sum is able to minimize the number of features to be computed for the next layer, and this requires less effort from the network to converge to a stable training.
	\item \textbf{Post-processing module}:
	it is an extra elaboration step applied to the merged features before eventually returning them.
	It permits the network to learn global features, jointly considering all the scales.
	To strengthen informative power of the final RoI, three module types have been considered for post-processing: a convolutional layer, a non-local layer \cite{wang2018non} and an attention layer \cite{zhu2019empirical}.
	Although the attention module is more complex because it requires also a fully-connected layer, our ablation analysis demonstrates that it is the best performing choice.
	The reason is that unlike the pre-processing module, the main objective of this layer is to eliminate useless information.
	In particular, the ``query content and relative position" configuration, called $\varepsilon_2$ in \cite{zhu2019empirical},  attention factor is used. This is more sensitive to the query content and have the higher impact on image contents.
\end{enumerate}

Summarizing, starting from a region produced by the RPN, for each scale, a fixed-size RoI is pooled from the region. The resulting $n$ feature maps are, first, separately pre-processed and, then, merged into a single feature map. Finally, post-processing is applied to extract global information. This architecture grants an equal contribution of each scale and benefits from the information embodied in all FPN layers by overcoming the limitations inherent in the arbitrary choice of a single FPN layer. It is worth noting that this procedure is valid for both object detection and instance segmentation.

%% file: experiments.tex
\section{Experiments}
\label{sec:experiments}

In this section two sets of experiments are reported.
The first set is a module-wise ablation analysis of the proposed GRoIE layer with the aim of finding the best combination of choices for each of the modules described in the previous section.
As was mentioned above, GRoIE can be plugged into architectures for both object detection (bounding box) and instance segmentation.

In the first set of experiments, we focus on object detection task only and employ the well-known Faster R-CNN as baseline.
In the second set, we apply GRoIE, with the best configuration found, to different architectures with the aim of showing the improvement in average precision for both object detection and instance segmentation.
This will allow us to show that the improvement produced by GRoIE is independent from both tasks as well as the utilized architecture.

\subsection{Dataset and Evaluation Metrics}
\label{subsec:dataset}

\noindent\textbf{Datasets}.
In order to evaluate our proposal, we performed experiments on MS COCO dataset 2017 \cite{lin2014microsoft} which is the \emph{de facto} standard dataset for large-scale object detection and instance segmentation tasks.
It is composed of 80 object categories and contains more than 116 thousand images in its training set.

\noindent\textbf{Evaluation Metrics}.
To extract the metrics, we used the official COCO python package.
The validation dataset, referred to as \emph{minival}, includes 5000 images.

The package calculates the Average Precision (AP) with different IoU (Intersection over the Union) thresholds for both bounding box and segmentation tasks.
The primary metric, indicated simply as $AP$, is calculated with IoU thresholds from 0.5 to 0.95.
Other metrics include $AP_{50}$ with the IoU threshold of 0.5 and $AP_{75}$ with 0.75.
In addition, separate metrics are calculated for small ($AP_s$), medium ($AP_m$) and large ($AP_l$) objects.

\subsection{Implementation details}

All the results with which we compare ours are not taken from the original papers, but they were obtained by training on the same hardware, with the same configuration (apart the RoI extractor) and by using the original authors' code when available.
These precautions are taken in order not to have the comparison affected by any small changes in either the configuration or the code.
We used MMDetection \cite{mmdetection} as base framework to develop our code.

The following base configuration was used for every experiment.
Experiments were conducted on 6 GPUs (Nvidia Tesla P100 with 12 GB of memory) for 12 epochs with an initial learning rate of 0.015, with a weight decay of 0.0001 after 9 and 11 epochs, a batch size of 2 images per GPU, and a random seed always equals to the number zero.
Since in most of the experiments reported in the literature, reference hardware is composed of 8 GPUs with batch size 2 and learning rate equal to 0.02, we followed the Linear Scaling Rule proposed in \cite{goyal2017accurate} to have a fair comparison.
The long edge and short edge of the images were resized to 1333 and 800, but the aspect ratio was maintained.
ResNet50 \cite{he2016deep} was used as backbone and RoI Align was selected for the RoI Pooling module (no ablation analysis was conducted on this module).

\subsection{Module-wise ablation analysis}
\label{Component ablation study}

In this section, we investigate how the choices of the GRoIE modules influence its final performance.
We compare our RoI Extractor architectures with the baseline represented by the single-layer RoI extractor proposed as part of the Faster R-CNN on \cite{ren2015faster} paper.

\begin{table}
  \begin{center}
    \begin{tabular}{l|c|c|c|c|c|c}
      Method & $AP$ & $AP_{50}$ & $AP_{75}$ & $AP_{s}$ & $AP_{m}$ & $AP_{l}$ \\
      \hline\hline
      baseline   & 36.5 & 58.4 & 39.1 & 21.9 & 40.4 & 46.8 \\
      sum        & \textbf{36.8} & \textbf{59.0} & \textbf{39.5} & \textbf{22.0} & \textbf{41.0} & \textbf{47.2} \\
      sum$+$     & 36.0 & 57.9 & 38.3 & 21.6 & 39.7 & 46.1 \\
      concat     & 36.1 & 58.0 & 38.6 & 21.4 & 40.3 & 45.7 \\
    \end{tabular}
  \end{center}
  \caption{Ablation analysis on aggregation module.}
  \label{ablation-aggregation}
\end{table}

\begin{figure}[t]
  \begin{center}
    \includegraphics[width=1\linewidth]{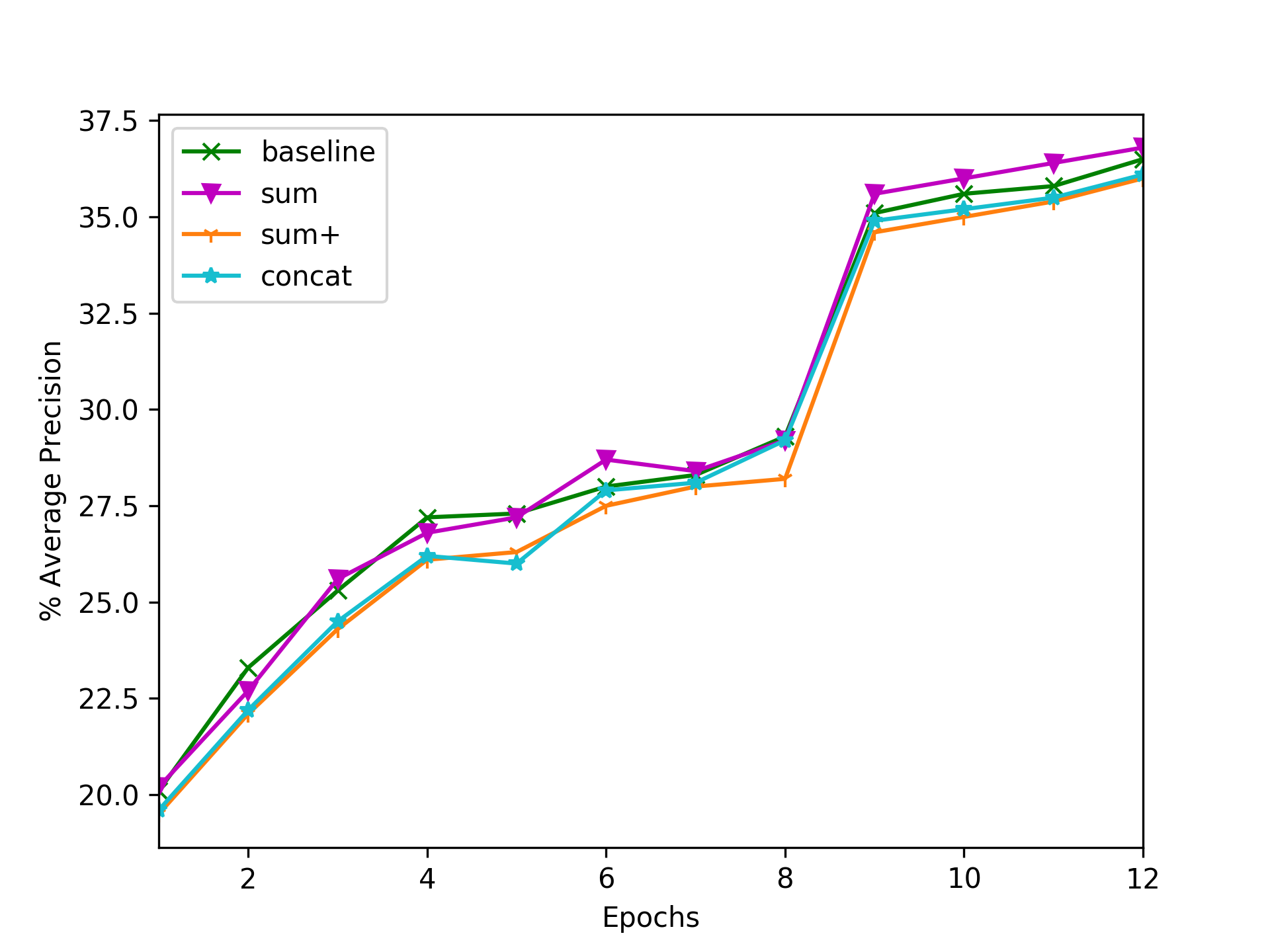}
  \end{center}
  \caption{Aggregation module analysis of average precisions trend on training.}
  \label{fig:ablation_aggregation}
\end{figure}

\noindent\textbf{Aggregation module analysis}.
We start from aggregation module because choosing how to merge the data technically has a significant importance on the architecture of the module itself.
In order to evaluate the effects of different choices separately, neither pre-processing nor post-processing are applied in this experiment.
Each FPN output layer is RoI pooled to create a 256 dimensional feature map and subsequently merged to form a single RoI.

There are mainly two choices for aggregating different branches: concatenation and summation.
In the first case, we need to reduce the feature maps from 1024 to 256 dimensions because we have 4 FPN layers, each one composed by 256 dimensions feature maps.
This can be easily done using a convolutional layer with 1x1 kernel.
A sum-based aggregation is simpler, but a fair comparison with concatenation is needed.
Therefore, in addition to a naive $sum$ operator, we included a variant of sum aggregation followed by a convolutional layer with 1x1 kernel as post-processing.
We call this $sum+$.

Table \ref{ablation-aggregation} shows the comparison between the proposed choices and a single-layer RoI extractor module (indicated as \dquote{baseline}).
To better justify our final choice, we show in Fig. \ref{fig:ablation_aggregation} the trend in average precision when the training epochs progress.
Looking at the results of $sum+$ and concatenation, one might argue that the integration of different FPN layers, the basis of our work, is not always beneficial.
This can be attributed to the added complexity which can be, in some cases, counterproductive and generate side effects.
In the case of $sum$, while at the beginning the trend is very similar, later in the training this operator achieves better accuracy with a stable trend, suggesting that this gap could potentially increase with more training epochs.
Therefore, we selected $sum$ operator for the aggregation module of GRoIE.


\begin{table}
  \begin{center}
    \begin{tabular}{l|c|c|c|c|c|c}
      Method & $AP$ & $AP_{50}$ & $AP_{75}$ & $AP_{s}$ & $AP_{m}$ & $AP_{l}$ \\
      \hline\hline
      baseline    & 36.5 & 58.4 & 39.1 & 21.9 & 40.4 & 46.8 \\
      conv 1x1    & 36.2 & 58.1 & 38.9 & 21.3 & 40.0 & 46.1 \\
      conv 3x3    & 37.0 & 58.6 & 40.1 & \textbf{22.0} & 40.9 & 47.0 \\
      conv 5x5    & \textbf{37.2} & \textbf{59.0} & \textbf{40.4} & 21.9 & \textbf{41.1} & \textbf{48.3} \\
      Non-local   & 36.5 & 58.5 & 39.0 & 21.9 & 40.5 & 46.7 \\
      Attention   & 36.4 & 58.3 & 39.1 & 21.8 & 40.4 & 46.7 \\
    \end{tabular}
  \end{center}
  \caption{Ablation analysis on pre-processing module.}
  \label{ablation-preprocessing}
\end{table}

\noindent\textbf{Pre-processing module analysis}.
For this ablation analysis, as mentioned above and based on the findings of the previous module, we chose the $sum$ operator for the aggregation module and did not apply any post-processing.
With regard to pre-processing, we consider three possible choices: using a convolutional layer with different kernel sizes, using a non-local module or using an attention module which was described in the previous section.

Table \ref{ablation-preprocessing} shows the comparison of these choices with the baseline as in the case of the aggregation module.
Regarding the convolutional layer, it can be noticed that by increasing the kernel size, the results are consistently improved.
This confirms the close correlation between neighboring features.
We should mention that the processed feature maps are only 7x7 in size.
This stopped us from increasing the kernel furthermore.


\begin{table}
  \begin{center}
    \begin{tabular}{l|c|c|c|c|c|c}
      Method & $AP$ & $AP_{50}$ & $AP_{75}$ & $AP_{s}$ & $AP_{m}$ & $AP_{l}$ \\
      \hline\hline
      baseline  & 36.5 & 58.4 & 39.1 & \textbf{21.9} & 40.4 & 46.8 \\
      conv 1x1  & 36.0 & 57.9 & 38.3 & 21.6 & 39.7 & 46.1 \\
      conv 3x3  & 36.6 & 58.3 & 39.3 & 21.3 & 40.5 & 46.6 \\
      conv 5x5  & 36.6 & 58.4 & 39.5 & 21.6 & 40.5 & 46.9 \\
      Non-local & 36.7 & \textbf{58.8} & 38.9 & 21.8 & \textbf{40.9} & 46.8 \\
      Attention & \textbf{36.8} & \textbf{58.8} & \textbf{39.9} & \textbf{21.9} & 40.4 & \textbf{47.0} \\
    \end{tabular}
  \end{center}
  \caption{Ablation analysis on post-processing module.}
  \label{ablation-postprocessing}
\end{table}

\noindent\textbf{Post-processing module analysis}.
Finally, we analyze the post-processing module, by keeping the $sum$ operator as aggregation strategy and not applying pre-processing.

Comparing Tables \ref{ablation-preprocessing} and \ref{ablation-postprocessing} which contain results for pre- and post-processing modules reveals a major difference.
While in the former, convolutional layers with different kernel sizes improve the results but non-local/attention modules do not, in the latter table the outcomes are opposite;
that is, improvement of convolutional layers is negligible, while non-local and attention methods bring about noticeable enhancement.
This can be explained by the fact that while in pre-processing there is the need to extract spatial contributions of the different layers where convolution acts correctly, in the post-processing phase the layers have already been merged by the aggregation module.
Therefore, convolution does not add significant information.
On the contrary, in post-processing, non-local and attention methods are able to remove useless information by focusing only on the significant parts of the image with attention mechanism.

%% file: experiments_sota.tex
\begin{table*}
  \begin{center}
    \begin{tabular}{l|c||c|c|c|c|c|c||c|c|c|c|c|c}
      & & \multicolumn{6}{c||}{Object detection} & \multicolumn{6}{c}{Instance segmentation} \\
      Method & Backbone & $AP$ & $AP_{50}$ & $AP_{75}$  & $AP_{s}$ & $AP_{m}$ & $AP_{l}$  & AP & $AP_{50}$ & $AP_{75}$  & $AP_{s}$ & $AP_{m}$ & $AP_{l}$\\
      \hline\hline
      Faster R-CNN  & r50-FPN & 36.5 & 58.4 & 39.1 & 21.9 & 40.4 & 46.8 & N/A & N/A & N/A & N/A & N/A & N/A \\
      +GRoIE (ours) & r50-FPN & \textbf{37.5} & \textbf{59.2} & \textbf{40.6} & \textbf{22.3} & \textbf{41.5} & \textbf{47.8} & N/A & N/A & N/A & N/A & N/A & N/A \\
      \hline\hline
      Grid R-CNN    & r50-FPN & 39.1 & 57.2 & 42.2 & 22.1 & 43.0 & 50.6 & N/A & N/A & N/A & N/A & N/A & N/A \\
      +GRoIE (ours) & r50-FPN & \textbf{39.8} & \textbf{58.1} & \textbf{42.9} & \textbf{23.6} & \textbf{43.9} & \textbf{51.5} & N/A & N/A & N/A & N/A & N/A & N/A \\
      \hline\hline
      Mask R-CNN    & r50-FPN & 37.3 & 58.9 & 40.4 & 21.7 & 41.1 & 48.2 & 34.1 & 55.5 & 36.1 & 18.0 & 37.6 & 46.7 \\
      +GRoIE (ours) & r50-FPN & \textbf{38.4} & \textbf{59.9} & \textbf{41.7} & \textbf{22.9} & \textbf{42.1} & \textbf{49.7} & \textbf{35.8} & \textbf{57.1} & \textbf{38.0} & \textbf{19.1} & \textbf{39.0} & \textbf{48.7} \\
      \hline\hline
      GC-net        & r50-FPN & 39.5 & 62.0 & 42.7 & \textbf{24.6} & 43.2 & 51.6 & 35.9 & 58.5 & 38.0 & \textbf{20.4} & 39.4 & 49.0 \\
      +GRoIE (ours) & r50-FPN & \textbf{40.3} & \textbf{62.4} & \textbf{44.0} & 24.2 & \textbf{44.4} & \textbf{52.5} & \textbf{37.2} & \textbf{59.3} & \textbf{39.8} & 20.2 & \textbf{41.0} & \textbf{51.2} \\
    \end{tabular}
  \end{center}
  \caption{Average precision w/ and w/o our GRoIE module.
  	       In the case of object detection networks, since they do not make image segmentation, an N/A has been inserted.}
  \label{sota-coco-bbox-mask-results}
\end{table*}

\begin{figure}[t]
  \begin{center}
    \includegraphics[width=1\linewidth]{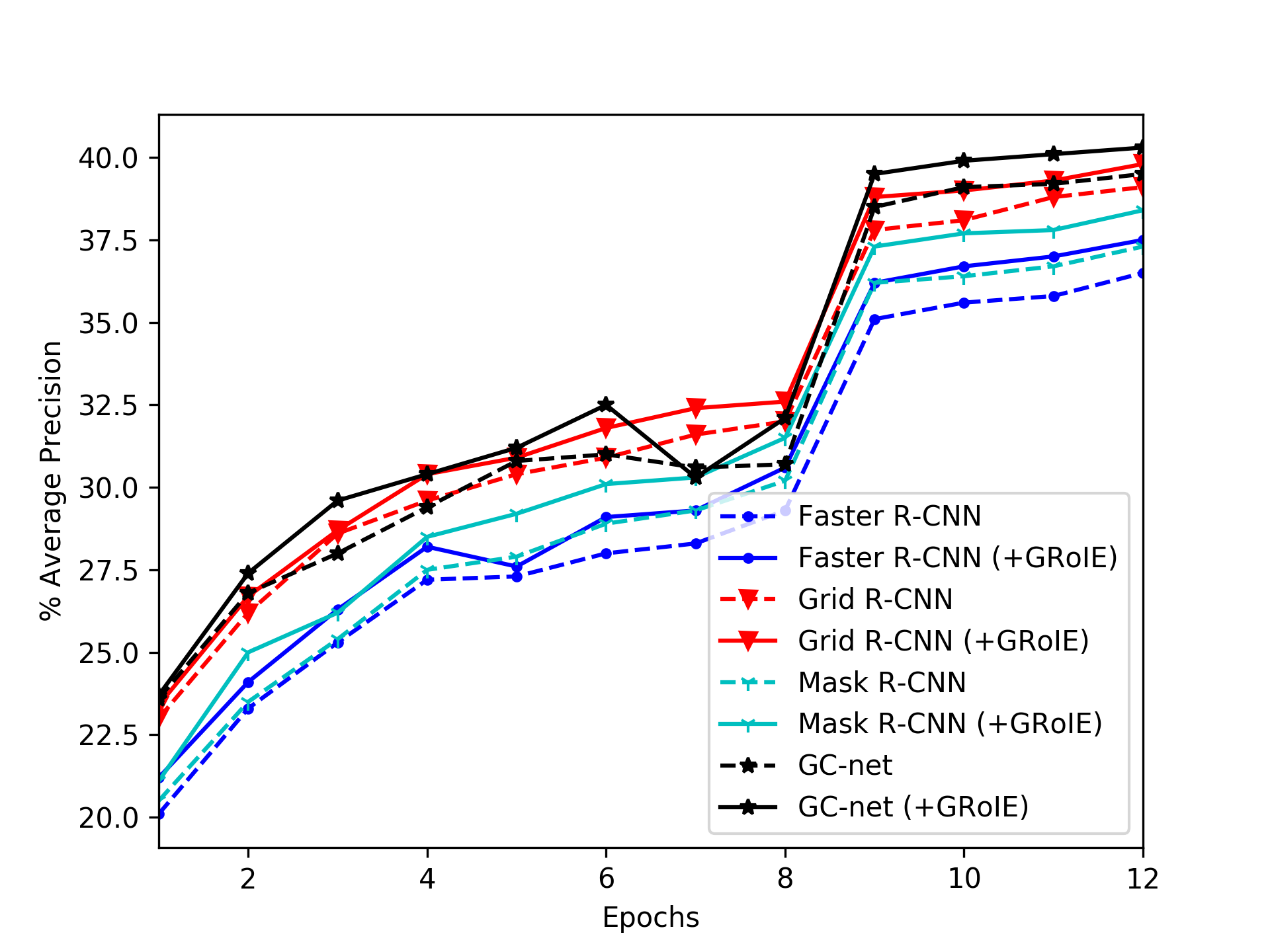}
  \end{center}
  \caption{Object detection average precision on \textit{minival} COCO dataset.}
  \label{fig:mask_best_ap_bbox}
\end{figure}

\begin{figure}[t]
  \begin{center}
    \includegraphics[width=1\linewidth]{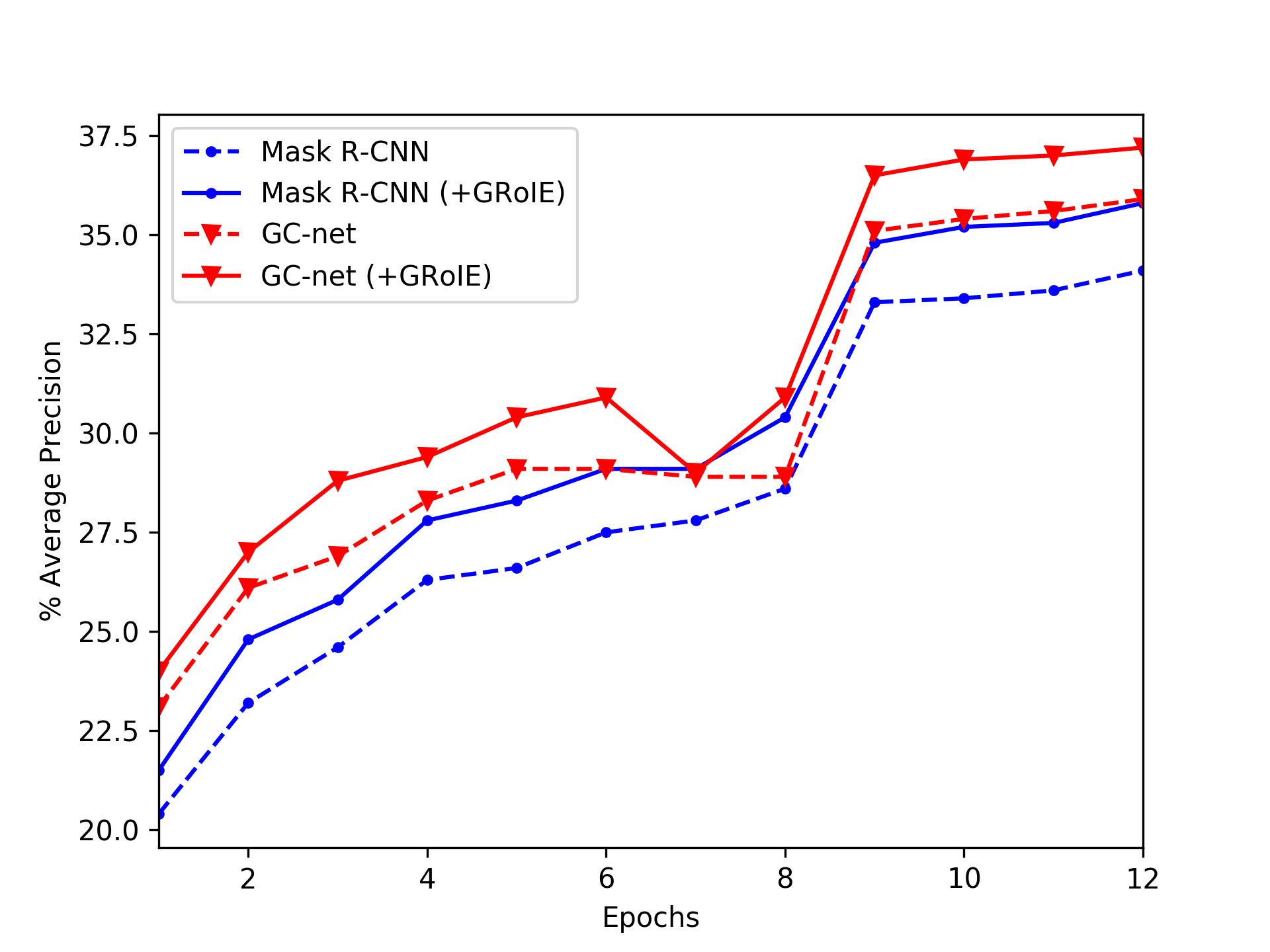}
  \end{center}
  \caption{Instance segmentation average precision on \textit{minival} COCO dataset.}
  \label{fig:mask_best_ap_segm}
\end{figure}

\subsection{Application of GRoIE to different architectures}
\label{Model ablation study}
As stated at the beginning of this section, the second set of experiments starts from the choices made on GRoIE modules based on the ablation analysis and integrates our proposed layer within several state-of-the-art architectures, with the aim of evaluating its benefits for both object detection and instance segmentation.
We have considered, first of all, the networks that best represent the two-stage networks: Faster R-CNN and Mask R-CNN.
Furthermore, we have taken into consideration the networks that have shown the best results in the recent years: Grid R-CNN \cite{lu2019grid} for object detection and GC-net \cite{cao2019gcnet} for instance segmentation too.
For the latter network, there are two RoI extractors. 
The first one is used for the detection part to extract the RoIs provided by the RPN;
the second one is used by the segmentation part to extract the RoIs provided by the detection.

For this experiment, we have thus replaced only the standard RoI extraction modules with GRoIE in its most performing configuration: $sum$ as aggregation function, 5x5 convolution for pre-processing and attention module for post-processing.

Table \ref{sota-coco-bbox-mask-results} shows the achieved results for both object detection (bounding boxes) and instance segmentation.
It is rather evident that the introduction of GRoIE as RoI extraction layer strongly contributes to an improvement in precision in all the tested architectures.
As expected, the amount of this improvement is not always the same and varies from a minimum of 0.7\% AP to a maximum of 1.1\% AP for bounding boxes, and from a minimum of 1.3\% AP to a maximum of 1.7\% AP for instance segmentation.
Looking at the other evaluation metrics, the gain is even more noticeable, with a maximum of 2.2\% for $AP_l$ in GC-net.

This improvement is even more evident from Figs. \ref{fig:mask_best_ap_bbox} and \ref{fig:mask_best_ap_segm}, where the average precision is illustrated with the progress of training epochs.
In these graphs, it can be seen that in later epochs the positive effect of GRoIE increases, suggesting that it can arguably be even higher with more training epochs.

%% file: conclusion.tex
\section{Conclusion}

In this paper, we proposed a novel RoI extraction layer for two-step architectures designed for object detection and instance segmentation.
The intuition underlying our proposal is that all the feature scales obtained by an FPN are potentially equally-useful for obtaining good final results.
The proposed layer, called GRoIE (Generic RoI Extractor), builds upon this intuition by first pre-processing each single layer, then aggregating them together, and finally applying attentive mechanisms as post-processing in order to remove useless (global) information.

Experiments are conducted on COCO dataset and a comprehensive ablation study has been conducted in order to select the best configuration of modules.
Furthermore, the addition of GRoIE to state-of-the-art two-step architectures for both object detection and instance segmentation has shown a consistent improvement in average precision in all the experiments.

While preliminary, the results reported in this paper are quite promising and seem to indicate the potentiality of GRoIE as novel extraction layer. 
As a consequence, our future works will concentrate on exploiting the modularity of GRoIE to further enhance the quality of the output features to improve the overall accuracy of different computer vision applications.
In addition, neural networks are now increasingly heavy to perform.
For this reason, an important field of exploration also for GRoIE regards precisely adopting every possible stratagem to lighten the workload while keeping performance unchanged.